# Hybrid Pointer Networks for Traveling Salesman Problems Optimization

**Ahmed Stohy[1], Heba-Tullah Abdelhakam[1], Sayed Ali[1], Mohammed Elhenawy[2], Abdallah A Hassan[1], Mahmoud Masoud[2*], Sebastien Glaser[2] and Andry Rakotonirainy[2]**

[1] Department of Computer and Systems Engineering, Minya University, Egypt
[2] Centre for Accident Research and Road Safety, Queensland University of Technology, Australia

Corresponding author: Mahmoud Masoud (Mahmoud.masoud@qut.edu.au).

**ABSTRACT** In this work, a novel idea is presented for combinatorial optimization problems, a hybrid network, which results in a superior outcome. We applied this method to graph pointer networks [1] expanding its capabilities to a higher level. We proposed a hybrid pointer network (HPN) to solve the travelling salesman problem trained by reinforcement learning. Furthermore, HPN builds upon graph pointer networks which is an extension of pointer networks with an additional graph embedding layer. HPN outperforms the graph pointer network in solution quality due to the hybrid encoder, which provides our model with a verity encoding type, allowing our model to converge to a better policy. Our network significantly outperforms the original graph pointer network for small and large-scale problems increasing its performance for TSP50 from 5.959 to 5.706 without utilizing 2opt, Pointer networks, Attention model, and a wide range of models, producing results comparable to highly tuned and specialized algorithms. We make our data, models, and code publicly available https://github.com/AhmedStohy/Hybrid-Pointer-Networks.

## I. INTRODUCTION

Due to the many applications of the travelling salesman problem (TSP) in many areas, it has received significant attention from the machine learning community in the past years. However, the developed neural combinatorial optimization models are still in the infantry stage. Generalization is still an unresolved problem when it comes to dealing with a big number of points with high precision.

The travelling salesman problem (TSP) is considered as one of the most significant and practical problems. Consider a salesman travelling to several areas, the salesman must visit each city just once while minimizing the total travel time. TSP is an NP-hard problem [2], which addresses the challenge of finding optimal solution in polynomial time.

Many approximation algorithms and heuristics, such as Christofides algorithm [3], local search [4], and the Lin-Kernighan heuristic (LKH) [5] have been developed to overcome the complexity of the exact algorithms which are guaranteed to yield an optimal solution but are frequently too computationally costly to be utilized in practice [6]. The pointer network [7], a seq2seq model [8], shows great potential for approximation solutions to combinatorial optimization problems such as identifying the convex hull and the TSP. It uses LSTM [9] as the encoder and an attention mechanism [10] as the decoder to extract features from city coordinates. It then predicts a policy outlining the next likely city by selecting a permutation of visited cities. The pointer network model is trained using the Actor-Critic technique [11].

Moreover, attention model [12] and [13] influenced by the Transformer architecture [10] tried to address routing difficulties such as the TSP and VRP. Graph pointer networks extended the traditional pointer networks with an additional layer of graph embedding, this transformation achieved a better generalization for a large-scale problem, but the GPN model without 2-opt still struggling for finding the optimal solutions for small and large scale.

The work proposed in this paper begins with how the performance of graph pointer networks can be improved without changing much of the architecture; an extra encoder layer is added alongside the graph embedding layer to act as a hybrid encoder, and this gives the model the ability to achieve good results; this will be discussed in greater detail in the HPN section.

Extensive results show that the proposed technique significantly outperforms previous DL-based methods on

TSP. The learnt model is more successful than typical hand-crafted rules in guiding the improvement process, and they may be further strengthened by simple ensemble methods. Furthermore, HPN generalize rather well to a variety of problem sizes, starting solutions, and even real-world datasets. It should be noted that the goal is not to outperform highly optimized and specialized traditional solvers, but to present a generalized model that can automatically learn good search heuristics on different problem sizes, which has a great value when applied to real-world problems.

## II. Travelling salesman problem (TSP)

TSP is a classic example of a combinatorial optimization problem that has been used in data clustering, genome sequencing, as well as other fields. TSP problem is NP-hard, and several exact, heuristic, and approximation algorithms have been developed to solve it. In this paper, TSP problems are assumed to be symmetric. The symmetric TSP is regarded as an undirected graph.

### 1. Asymmetric vs. symmetric TSP

The distance between two cities in the symmetric TSP is the same in each opposite direction, producing an undirected network. This symmetry cuts the number of alternative solutions in half. Paths may not exist in both directions in the asymmetric TSP, or the distances may be different, resulting in a directed graph.

### 2. Directed vs. undirected graphs

Edges in *undirected graphs* do not have a direction. Each edge may be travelled in both directions, indicating a two-way connection. The edges of *directed graphs* have a direction. The edges represent a one-way connection, as each edge may only be travelled in one direction.

A full undirected graph can be defined as $C = (V, E)$ where $V$ is the vector of vertices of the graph $C$, and $E$ is the vector of edges between these vertices. In this study, the TSP's graph is complete so every node has an edge to each of the other vertices in the graph.

$$C \text{ is symmetric if } (\forall i, j : e_{ij} = e_{ji}),$$

In the context of this paper, $e_{ij}$ equals the distance between the vertices $i$ and $j$. Given a set $V$ of cities $n$ in a two-dimensional space, the objective is to find the optimal Hamiltonian path that minimizes the total tour length [14]:

$$L(\pi | V) = \left\| v_{\pi(n)} - v_{\pi(1)} \right\|_2 + \sum_{i=1}^{n-1} \left\| v_{\pi(i)} - v_{\pi(i+1)} \right\|_2 \quad (0.1)$$

Where $\left\| . \right\|_2$ is $\ell_2$ norm and $\pi$ denote as a tour.

## III. REINFORCMENT LEARNING (RL)

Reinforcement learning (RL) is the process of learning what to perform to increase a numerical reward signal. The agent isn't instructed which actions to perform but must experiment to determine which acts offer the greatest expected reward. To begin, we will define the notation used to represent the TSP as a reinforcement learning problem. Let $S$ be the state space and $A$ the action space. Each state $s_t \in S$ is defined as the set of all previously visited cities. The action $a_t \in A$ is defined as the next selected city from the set of possible cities, our model is considered a sequential one that given an instance $a_t$ (selected input city) outputs a probability distribution over the next candidates from the remaining cities that have not been chosen. We can define our policy as:

$$P_\theta(\pi | x) = P[A_t = a | S_t = s], \quad (0.2)$$

From which we can sample to obtain a tour π. In order to train our model, we define the loss [12]:

$$L(\theta | s) = E_{P_\theta(\pi|x)}[L(\pi)], \quad (0.3)$$

Where $L(\pi)$ is the cost of the tour that we are attempting to minimize. Recall the REINFORCE's [15] equation with baseline which is an extension from policy gradient algorithm [16]:

$$\nabla J \approx E[(L(\pi) - b(s))\nabla \log \pi(a | s)]. \quad (0.4)$$

Where $b(s)$ is the baseline subtracted from the cost to eliminate the policy gradient variance. The optimal baseline is one that lowers variation as much as possible while simultaneously speeding up the training process. As a result, we employ the approach given by [12]:

**Algorithm 1. REINFORCE with Rollout Baseline [12]**

| | | |
|---|---|---|
| 1 | : | input: number of epochs E, steps per epoch T, batch size B, significance α |
| 2 | : | init θ, $\theta^{BL} \leftarrow \theta$ |
| 3 | : | for epoch = 1,..., E do |
| 4 | : |    for step = 1,..., T do |
| 5 | : |       $s_i \leftarrow$ RandomInstance () ∀ i ∈ {1,..., B}v |
| 6 | : |       $\pi_i \leftarrow$ SampleRollout ($s_i, p_\theta$) ∀ i ∈ {1,..., B} |
| 7 | : |       $\pi_i^{BL} \leftarrow$ GreedyRollout ($s_i, p_\theta^{BL}$) ∀ i ∈ {1,..., B} |
| 8 | : |       $\nabla L \leftarrow \sum_{i=1}^{B} L(\pi_i) - L(\pi_i^{BL}) \nabla_\theta \log p_\theta(\pi_i)$ |
| 9 | : |       θ ← Adam (θ, ∇L) |
| 10 | : |    end for |
| 11 | : |    if OneSidedPairedTTest ($p_\theta, p_\theta^{BL}$) < α then |

| 12 | : | $\theta^{BL} \leftarrow \theta$ |
| 13 | : | end if |
| 14 | : | end for |

## IV. Hybrid Pointer Network (HPN)

HPN is inspired by the Graph pointer network (GPN). GPN is a modified variant of the classic pointer network (PN).

Graph pointer networks have been used to tackle TSP. Building on this approach, in this paper:

- The graph embedding layer is combined with the transformer's encoder to produce multiple embeddings for the feature context.
- An extra decoder layer is added to operate as a multi-decoder structure network to improve the agent's decision-making process throughout the learning phase.
- Finally, we switch our learning algorithm from a central self-critic [17] to an actor-critic one as suggested by Kool [12].

An additional graph embedding layer is added above the pointer network, allowing the model to figure out the complicated relationships between graph nodes in large-scale problems. However, it still struggles to find a globally optimal strategy for small and large TSP problems. This study proposes extending the network architecture to converge to a better policy for small, medium, and large sizes. The proposed HPN is shown in Figure 1. HPN consists of a mixture of several encoder's architecture and multi decoder based on the attention concept.

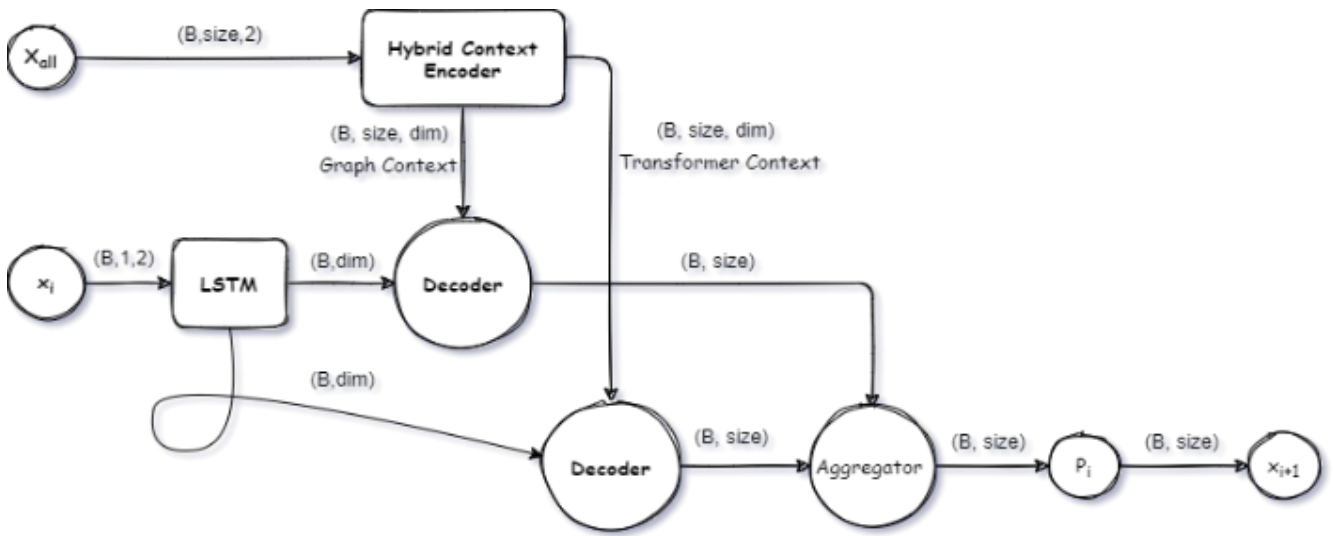

*Figure 1. Architecture of HPN which combining a hybrid context encoder with a multi-attention decoder.*

### 1. Hybrid Encoder

As illustrated in Figure 2, the proposed encoder consists of two parts: the hybrid context encoder, which encodes the Feature vector into two contextual vectors and the point encoder, which encodes the currently selected city by LSTM. Two different encoders are employed for the hybrid context encoder. The first encoder is a typical transformer encoder with multi-head attention and residual connection with batch normalizing layer, the transformer's encoder equations with a single head are [18]:

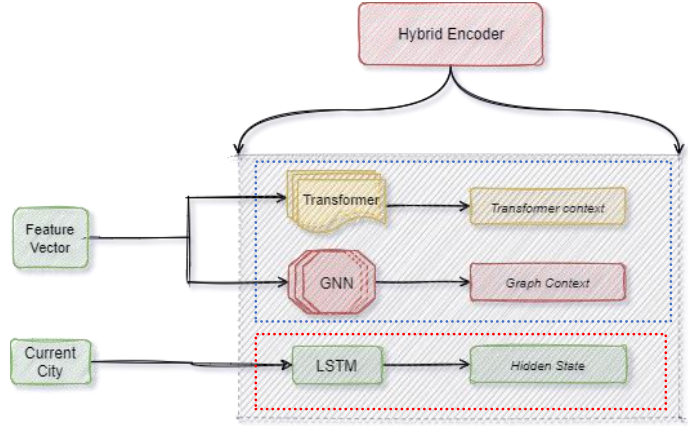

*Figure 2. Hybrid encoder consists of Transformer's encoder and Graph embedding layer as a hybrid context encoder (blue dotted box) and the point encoder (red dotted box) for the current city.*

$$H^{enc} = H^{l=L^{enc}} \in R^{(n+1)\times d} \quad (0.5)$$

$$H^l = softmax\left(\frac{Q^l K^{l^T}}{\sqrt{d}}\right) V^l \in R^{(n+1)\times d}, \quad (0.6)$$

$$Q^l = H^l W_Q^L \in R^{(n+1)\times d}, W_Q^l \in R^{d\times d}, \quad (0.7)$$

$$K^l = H^l W_K^L \in R^{(n+1)\times d}, W_K^l \in R^{d\times d}, \quad (0.8)$$

$$V^l = H^l W_V^L \in R^{(n+1)\times d}, W_V^l \in R^{d\times d} \quad (0.9)$$

Where $W_Q^L, W_K^L$ and $W_V^L$ are learnable parameters, $H^{enc}$ is a matrix contains the encoded nodes, $Q^l, K^l$ and $V^l$ are a query, key and value of the self-attention.

The second one is the graph embedding layer. The graph embedding layer context is acquired by directly encoding the context vector obtained from coordinates of cities. Because we are only considering symmetric TSP, the graph is full. As a result, the graph embedding layer can be written as [1]:

$$X^l = \gamma\, X^{l-1} W_g + (1-\gamma)\, \varphi_\theta\left(\frac{X^{l-1}}{|N(i)|}\right) \quad (0.10)$$

Where $X^l \in R^{N\times d_l}$, and $\varphi_\theta : R^{N\times d_{l-1}} \to R^{N\times d_l}$ is the aggregation function, $\gamma$ is a trainable parameter, $W_g \in R^{d_{l-1}\times d_l}$ is trainable weight matrix and $N(i)$ the adjacency set of node i.

For the point encoder which encodes the current selected city, each city coordinates $x_i$ $x_i$ is embedded into a higher dimensional vector $x \in R^d$, where d is the hidden dimension. The vector $x$ for the current city $x_i$ is then encoded by an LSTM. The hidden variable $x_i^h$ of the LSTM is passed to both the decoder of the current stamp and the encoder of the next time stamp.

### 2. Multi-decoder

To begin the decoding phase, a placeholder is added for the first iteration of the decoding to select the best location to start the tour, the decoder is based on the attention mechanism of a pointer network and outputs the pointer vector $u_i$, which is then sent through a Softmax layer to build a distribution across the following candidate cities. The attention mechanism and the pointer vector $u_i$ are defined as follows [1]:

$$u_i^{(j)} = \begin{cases} V^T \cdot \tanh\left(W_r r_j + W_q q\right) & \text{if } j \neq \sigma(k), \forall k < j, \\ -\infty & \text{otherwise,} \end{cases}$$

Where $u_i^{(j)}$ is the j-th entry of the vector $u_i$, $W_r$ and $W_q$ are trainable parameters, q is the query vector from the hidden state of the LSTM, is a reference vector containing the contextual information from all cities.

The encoded context from the transformer's encoder is used as a reference for the first decoder layer and the context obtained from the graph embedding layer is used as the reference for the second decoder layer, as illustrated in
Figure 1.

For determining the distribution policy across the candidate cities, four different operations can be employed for the aggregator:

- The first option is to add the two attention vectors from each decoder layer, which are provided by:
$$\pi_\theta(a_i | s_i) = p_i = softmax\,(u_i^1 + u_i^2)$$
- The second option is to take the maximum value between these two vectors, which is indicated as follows:
$$\pi_\theta(a_i | s_i) = p_i = softmax(\max(u_i^1, u_i^2))$$
- The third option is to take the mean as follows:
$$\pi_\theta(a_i | s_i) = p_i = softmax\,(average(u_i^1, u_i^2))$$
- The final option is to concatenate both of them and feed the concatenated vector into a single embedding

layer, letting the model to decide how to aggregate them; we can describe this notion as follows:

$$\pi_\theta(a_i \mid s_i) = p_i = softmax(\vartheta_\theta(cat(u_i^1, u_i^2))).$$

Where $\vartheta_\theta : R^{N \times 2} \to R^{N \times 1}$ is the aggregation function. In the result section, we displayed the outcome for each one of them.

## V. EXPERIMENTS

In our experiments. The training data is generated randomly from a $[0,1]^2$ uniform distribution. In each epoch, the training data is generated on the fly. The hyperparameters provided in *Table 1* are used in the following experiments.

*Table 1*
*Hyperparameters used for training*

| Parameter | Value | Parameter | Value |
|---|---|---|---|
| Graph Embedding | 3 | Learning rate | 1e-4 |
| Transformer Encoder | 6 | Batch size | 512 |
| Feed-forward dim | 512 | Training steps | 2500 |
| Optimizer | Adam | Tanh clipping | 10 |

### 1. Small-scale experiments

We begin our experiments with a difficult barrier: which aggregator function will assist our model in achieving better results? Indeed, it is difficult to answer this question without experimenting all of them; we examined the above-mentioned suggestions for this component and recorded the training performance results. Figure 3 illustrate these results.

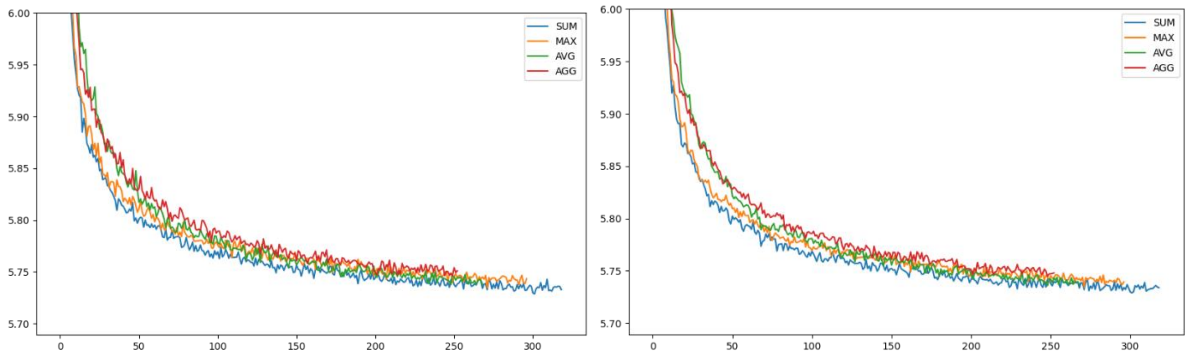

*Figure 3. Training performance for the actor (on Top) and the critic (on Bottom) where the total tour length on the y-axis and the number of epochs on x-axis indicating that when we apply the sum operation between the two attention's vectors, the model converges a little fast compared with the others*

We can conclude from the above figure that the summation has excellent performance at first, but by the middle of training, the average has caught it, the maximum and the single-layer aggregation have a little higher result, so we decide to stop examining them.

For tackling the small-scale. We use TSP50 instances to train our HPN model. TSP50's average training time for each epoch is 19 minutes while utilizing one instance of NVIDIA Tesla P100 GPU. We compare the performance of our model on small-scale TSP to earlier studies such as Graph pointer networks, the Attention Model, the pointer network, s2v-DQN [19], the Transformer Network [18] and other heuristics, e.g. 2-opt heuristics, Christofides algorithm and random insertion. The results are shown in Figure 4 which compares the approximate tour length to the optimal solution. Small number indicates a better result.

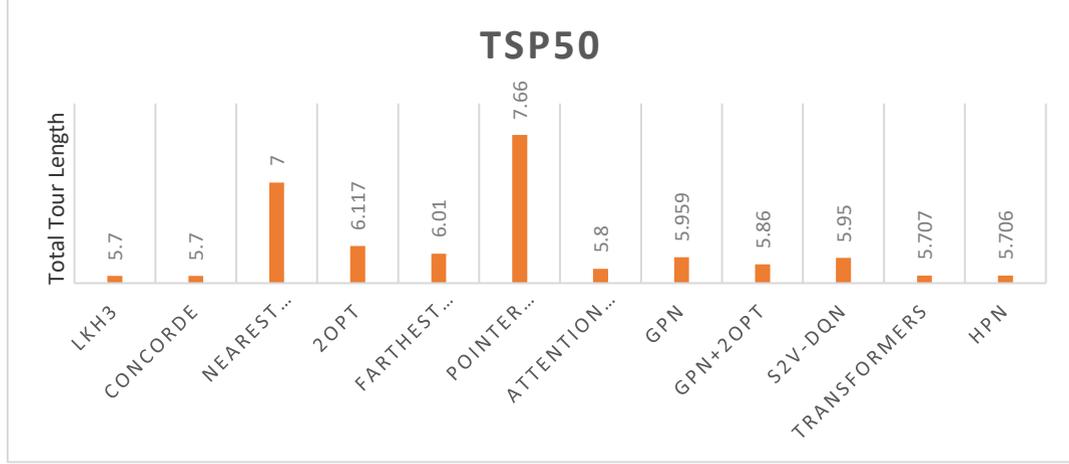

*Figure 4 Comparison of TSP50 results*

As demonstrated in Figure 4 our HPN model surpasses the current models and achieves the state-of-the-art solution for TSP50. Our model outperforms the graph pointer network by a wide margin, increasing its performance for TSP50 from 5.959 to 5.706 without utilizing 2opt, which is a huge success for the hybridization concept.

## 2. Large-scale experiments

For achieving the best possible generalization out of our model, instead of just using the cities' coordinates as a context for both the graph encoder and the transform encoder, we replace it with a feature context that accelerates the training convergence for the large-scale problems. The feature context includes the vector context previously used by [1] concatenated with the Euclidean distance, where the vector context is just a subtraction operation between the coordinates of the currently selected city with the others.

Our feature extractor component does this job as illustrated in Figure 5. It is essential in the proposed HPN model since it extracts the most relative information and feeds it to the encoder as a context. "Suppose that $X_i = [x_i^T, ..., x_i^T]^T \in R^{N \times 2}$ is a matrix with identical $N$ rows. We define $\overline{X}_t = X - X_i$ as the vector context. The j-th row of $X_i$ is a vector pointing from node $i$ to node $j$ and X is the matrix that contains coordinates'' of all cities. We expanded this notion by adding the Euclidean distance between the currently selected city and other cities to the vector context.

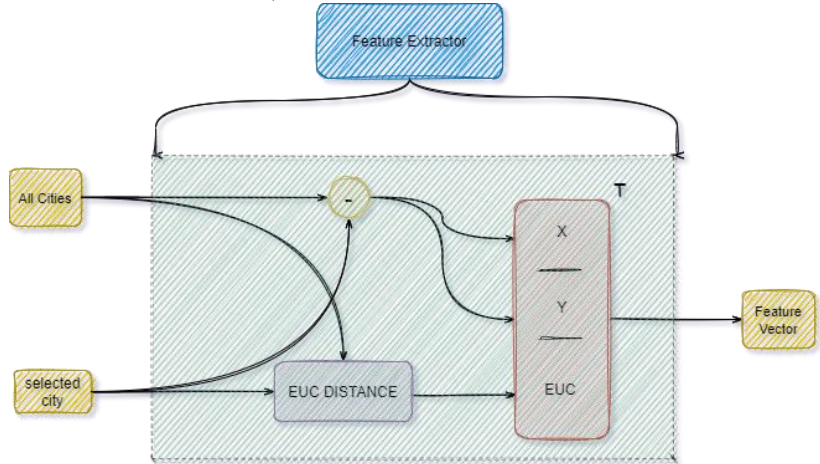

*Figure 5.* Architecture of feature extractor which combining both vector context with the Euclidian distance and output a Feature vector.

The Euclidean distance between two points $(x_i, y_i)$ and $(x_j, y_j)$ is shown in $(0.10)$:

$$d = \sqrt{(x_i, y_i)^2 + (x_j, y_j)^2} \quad (0.10)$$

Where x and y are the coordinates of each city.

Using the feature extractor, we train our large model in TSP50, validate with TSP500, 10 epochs, 1e-3 learning rate with leaning rate decay 0.96 and 100 for tanh clipping. Some sample tours are shown in Figure 6 in which we solve TSP50-250-500-100 with HPN+2opt.

Table 2 summarizes our result which showing that our model generalizes better than GPN. For the sake of a fair comparison with the state-of-the-art (i.e., GPN), we used 2opt local search technique to fine tune the HPN's tours for larger sizes of instances. As shown in Table 2 and Figure 7, our models outperform the GPN, GPN+2opt, PN, AM, and 2opt models. Moreover, HPN+2opt returns near optimal tours and generalizes better than the GPN on a large-scale instance.

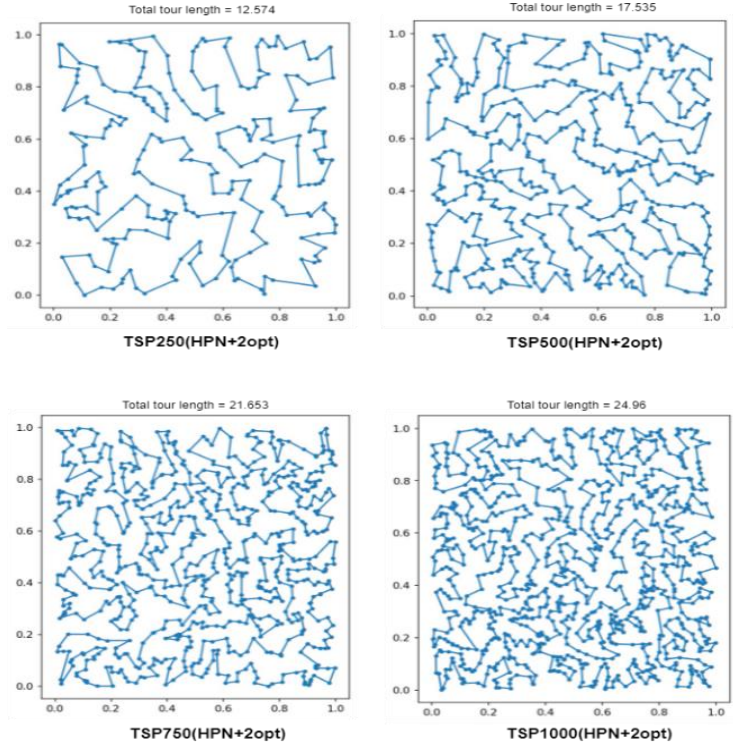

*Figure 6 Sample tours for TSP250-500-750-1000 solved by HPN+2-opt*

*Table 2.*
*TSP's result using Hybrid pointer network Model (HPN) vs baselines. Each result is obtained by averaging on 10k random TSP instances for TSP50 and 1K random instances for larger sizes.*

|  | TSP50 | | TSP250 | | TSP500 | | TSP750 | | TSP1000 | |
| Method | Obj. | Time | Obj. | Time | Obj. | Time | Obj. | Time | Obj. | Time |
| --- | --- | --- | --- | --- | --- | --- | --- | --- | --- | --- |
| LKH3 | 5.70 | 300s | 11.893 | 9792s | 16.542 | 23070s | 20.129 | 36840s | 23.130 | 50680s |
| Concorde | 5.70 | 120s | 11.89 | 1894s | 16:55 | 13902s | 20.10 | 32993s | 23.11 | 47804s |
| Nearest Neighbor | 7.00 | 0s | 14.928 | 25s | 20.791 | 60s | 25.219 | 115s | 28.973 | 136s |
| 2-opt | 6.117 | 7.92s | 13.253 | 303s | 18.600 | 1363s | 22.668 | 3296s | 26.111 | 6153s |
| Farthest Insertion | 6:01 | 2s | 13.026 | 33s | 18.288 | 160s | 22.342 | 454s | 25.741 | 945s |
| OR-Tools (Savings) | -- | -- | 12.652 | 5000s | 17.653 | 5000s | 22.933 | 5000s | 28.332 | 5000s |
| OR-Tools (Christofides) | -- | -- | 12.289 | 5000s | 17.449 | 5000s | 22.395 | 5000s | 26.477 | 5000s |
| Pointer Net | 7.66 | -- | 14.249 | 29s | 21.409 | 280s | 27.382 | 782s | 32.714 | 3133s |
| Attention Model | 5.80 | 2s | 14.032 | 2s | 24.789 | 14s | 28.281 | 42s | 34.055 | 136s |
| GPN | 5.959 | 1.75s | 13.679 | 32s | 19.605 | 111s | 24.337 | 232s | 28.471 | 393s |
| GPN+2opt | 5.867 | 6.5s | 12.942 | 214s | 18.358 | 974s | 22.541 | 2278s | 26.129 | 4410s |
| s2v-DQN | 5.95 | -- | 13.079 | 476s | 18.428 | 1508s | 22.550 | 3182s | 26.046 | 5600s |
| Transformers (Gr.) | 5.707 | 13.7s | 14.60 | 4s | 23.63 | 10s | 30.77 | 15s | -- | -- |
| HPN (Gr.) **ours** | 5.706 | 0.36s | 13.44 | 16s | 18.94 | 48s | 23.15 | 100s | 26.64 | 168s |
| HPN+2opt **ours** | -- | -- | 12.78 | 315s | 17.95 | 1460s | 21.95 | 3405s | 25.21 | 6480s |

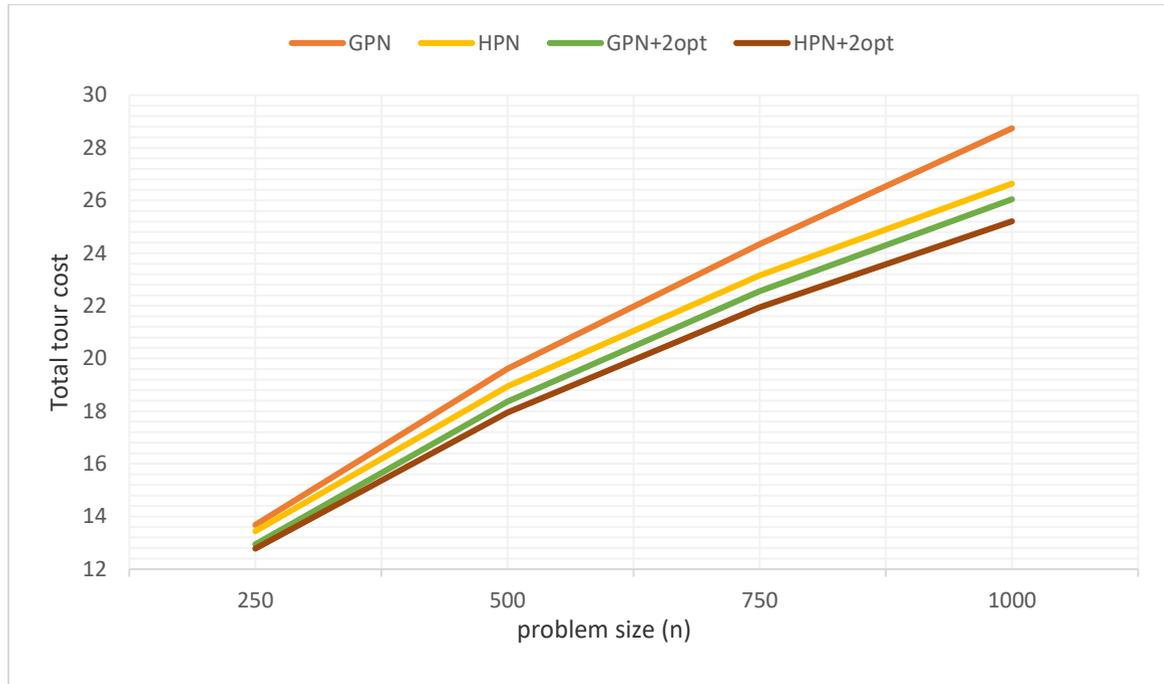

**Figure 7 Large-scale results from GPN, HPN, GPN+2opt, and HPN+2opt demonstrate that the gap between our model and GPN increases as the problem size increases.**

### 3. Benchmark instances results and statistical analysis

To validate our model against the standard benchmark instances, we employed varied-size examples from the public libraries TSPLIB and World TSP . The benchmark dataset consists of 34 instances. The naming convention of instances consists of the first few letters of the instance location and the problem size n. For example, the instance eg7146 has 7146 points in Egypt. The instance sizes vary from 400 to 10639 nodes(cites). The normalized and actual tour length in km and the testing time in seconds are reported in Table 3.

To understand how HPN is performing compared to the GPN (the state-of-the-art network) in terms of the tour cost and testing time shown in Table 3, we did statistical comparison between these two networks. The statistical model should consider the dependency between the observations shown in Table 3.

In other words, we should realize that the tour cost of HPN and GPN for the same instance are correlated. Moreover, the testing times of same instance using the two networks are correlated as well. Therefore, we used a generalized linear mixed effects model to explain the variability of the tour cost and testing time as a function of the network used and the size of the network. We used one indicator variables to code the GPN and HPN.

In Table 4, we compared the tour cost for the HPN and GPN. The p-value of the GPN indicator variable is <.0001 and we conclude that the tour cost of the GPN is statistically significantly higher than the tour cost of the HPN. However, as shown in Table 5, the testing time of the HPN is statistically significantly higher than the testing time of the GPN

*Table 3.*
*Evaluation on real world TSPLIB dataset using HPN and HPN+2opt.*

| Benchmark | HPN Obj. | HPN Time | HPN+2opt Obj. | HPN+2opt Time | Un-normalized tour length HPN+2opt in km | GPN Obj. | GPN Time | GPN+2opt Obj. | GPN+2opt Time | Un-normalized tour length GPN+2opt in km |
|---|---|---|---|---|---|---|---|---|---|---|
| rd400 | 17 | 2s | **16.6** | 2.3s | **16538** | 18 | 1s | 16.9 | 2.1s | 16827 |
| gr431 | 10.4 | 1.3s | **9.9** | 1.8s | **2473** | 11.3 | 1s | 10.3 | 2.3s | 2689 |
| d493 | 13.4 | 1.3s | **11.2** | 3.6s | **38404** | 12.9 | 1.1s | 11.7 | 2.3s | 40098 |
| att532 | 14.8 | 1.4s | **13.2** | 3.4s | **96916** | 14.9 | 1.1s | 13.5 | 3.4s | 1029393 |
| pa561 | 18 | 2.8s | **17.3** | 5.3s | **16357** | 19 | 1.3s | 17.9 | 2.8s | 16971 |
| u574 | 18.3 | 1.4s | **17.2** | 3.6s | **41323** | 22.7 | 1.6s | 17.5 | 3.4s | 42874 |
| d657 | 16.3 | 1.8s | **15.2** | 5s | **53921** | 17.8 | 1.6s | 15.8 | 4.1s | 55692 |
| gr666 | 14.5 | 1.7s | **13.5** | 5.3s | **3731** | 15.9 | 1.67s | 13.9 | 5.5s | 3840 |
| u724 | 22.5 | 2.8s | **21.1** | 5.4s | **48742** | 22.7 | 1.5s | 21.4 | 4.7s | 49105 |
| rat783 | 25.1 | 2.2s | **24.7** | 4.4s | **10887** | 26.9 | 1.9s | 25 | 5.1s | 11015 |
| dsj1000 | 18.8 | 3s | **17.2** | 12s | **20715222** | 20.1 | 2.1s | 17.5 | 11.6s | 21020136 |
| u1060 | 23 | 2.8s | **20** | 11.4s | **286109** | 23.6 | 2.3s | 21 | 11.1s | 289678 |
| d1291 | 19.6 | 4s | **16.6** | 17s | **57713** | 20.1 | 2.6s | 16.7 | 20.1s | 57964 |
| nrw1379 | 27.2 | 4s | **25.8** | 15s | **61163** | 30.4 | 3.1s | 27.1 | 19.8s | 64005 |
| u1432 | 33.9 | 3.8s | **32.8** | 14.53s | **166739** | 36.2 | 2.9s | 33.6 | 16.3s | 170875 |
| vm1748 | 28.8 | 5s | **24.9** | 38s | **386483** | 29.3 | 3.8s | 25.9 | 31.15s | 395392 |
| rl1889 | 27 | 6s | **23.2** | 39s | **365362** | 29.6 | 4s | 23.6 | 34.1s | 373652 |
| u2152 | 36.9 | 6s | **32.3** | 43s | **75754** | 38.7 | 4.4s | 32.7 | 42.5s | 77622 |
| pr2392 | 38 | 7s | **35.2** | 70s | **426199** | 42.8 | 5.3s | 36.2 | 57.4s | 436979 |
| pcb3038 | 46.1 | 10s | **43.** | 117s | **154445** | 52.7 | 6.5s | 44 | 105s | 156771 |
| nu3496 | 26.4 | 9s | **23.8** | 144s | **105314** | 32.6 | 8.1s | 24.3 | 154s | 107965 |
| fl3795 | 16.4 | 2s | **14.7** | 170s | **30786** | 23.7 | 8.4s | 14.5 | 258.7s | 30402 |
| fnl4461 | 29.9 | 15s | **46.7** | 220s | **203653** | 59 | 9.3s | 49 | 166.7s | 214703 |
| ca4663 | 26.2 | 14s | **23.8** | 248s | **1585498** | 37.2 | 10.1s | 24.3 | 273.7s | 1621202 |
| rl5934 | 43.9 | 22.6s | **37.9** | 501s | **630300** | 54.8 | 13s | 39.6 | 426.3s | 645335 |
| tz6117 | 46 | 16s | **40.3** | 431s | **434403** | 55.6 | 12.7s | 41.6 | 409.1s | 448478 |
| eg7146 | 21.7 | 19.3s | **19** | 440s | **187376** | 35.3 | 16.7s | 19.2 | 633.3s | 189608 |
| pla7397 | 51.8 | 26s | **42.6** | 906s | **25169496** | 71.1 | 16.3s | 44.5 | 598.4s | 26287346 |
| ym7663 | 32.8 | 21s | **30.3** | 594s | **281001** | 57.7 | 16.6s | 30.8 | 792.5s | 286558 |
| ei8246 | 58.3 | 22s | **45** | 585s | **226498** | 75.5 | 18.1s | 55.8 | 737.4s | 233699 |
| ar9152 | 40.9 | 26s | **36.9** | 834s | **972758** | 56.5 | 19.7s | 38.2 | 1009.3s | 1002739 |
| ja9847 | 28.6 | 28s | **24.4** | 1090s | **542887** | 41.5 | 21.4s | 24.7 | 1438.5s | 550364 |
| fi10639 | 56 | 49s | **51.5** | 1206s | **573948** | 81.6 | 24.2s | 52.8 | 1238.6s | 5888001 |

*Table 4:*
*the fixed effects coefficients of the model explaining the tour cost in terms of the HPN/GPN and the problem size*

| Name | Estimate | SE | tStat | DF | pValue | Lower | Upper |
|---|---|---|---|---|---|---|---|
| (Intercept) | 16.01394 | 2.47475 | 6.470932 | 65 | <.0001 | 11.07152 | 20.95636 |
| size | 0.003976 | 0.000526 | 7.557891 | 65 | <.0001 | 0.002926 | 0.005027 |
| GPN | 6.485472 | 1.229939 | 5.273005 | 65 | <.0001 | 4.029116 | 8.941829 |

*Table 5:*
*the fixed effects coefficients of the model explaining the testing time in terms of the HPN/GPN and the problem size*

| Name | Estimate | SE | tStat | DF | pValue | Lower | Upper |
|---|---|---|---|---|---|---|---|
| (Intercept) | 0.989527 | 0.649926 | 1.522523 | 65 | 0.132729 | -0.30846 | 2.287518 |
| size | 0.002742 | 0.000118 | 23.30437 | 65 | <.0001 | 0.002507 | 0.002977 |
| GPN | -2.79206 | 0.735072 | -3.79835 | 65 | <.0001 | -4.2601 | -1.32402 |

We repeat the same analysis for the tour cost and the testing time of the HPN+2opt and GPN+2opt. As shown in Table 6 and Table 7, the HPN+2opt has statistically significantly lower testing time and tour cost. This is expected because HPN returns a better tour as an initial point in the solution space which helps 2opt to find better final tour in less time. Finally, for the sake of completeness we visualized four constructed tours using HPN + 2opt in Figure 8.

*Table 6:*
*the fixed effects coefficients of the model explaining the tour cost in terms of the HPN+2opt/GPN+2opt and the problem size*

| Name | Estimate | SE | tStat | DF | pValue | Lower | Upper |
|---|---|---|---|---|---|---|---|
| (Intercept) | 17.96973 | 2.224075 | 8.079641 | 65 | <.0001 | 13.52795 | 22.41152 |
| size | 0.002559 | 0.000488 | 5.243273 | 65 | <.0001 | 0.001584 | 0.003534 |
| GPN | 0.724771 | 0.097485 | 7.434709 | 65 | <.0001 | 0.53008 | 0.919461 |

*Table 7:*
*the fixed effects coefficients of the model explaining the testing time in terms of the HPN+2opt/GPN+2opt and the problem size*

| Name | Estimate | SE | tStat | DF | pValue | Lower | Upper |
|---|---|---|---|---|---|---|---|
| (Intercept) | -143.885 | 25.28218 | -5.69116 | 65 | 3.26E-07 | -194.377 | -93.393 |
| size | 0.106226 | 0.005288 | 20.08878 | 65 | <.0001 | 0.095665 | 0.116786 |
| GPN | 42.49765 | 15.33099 | 2.77201 | 65 | <.0001 | 11.87955 | 73.11574 |

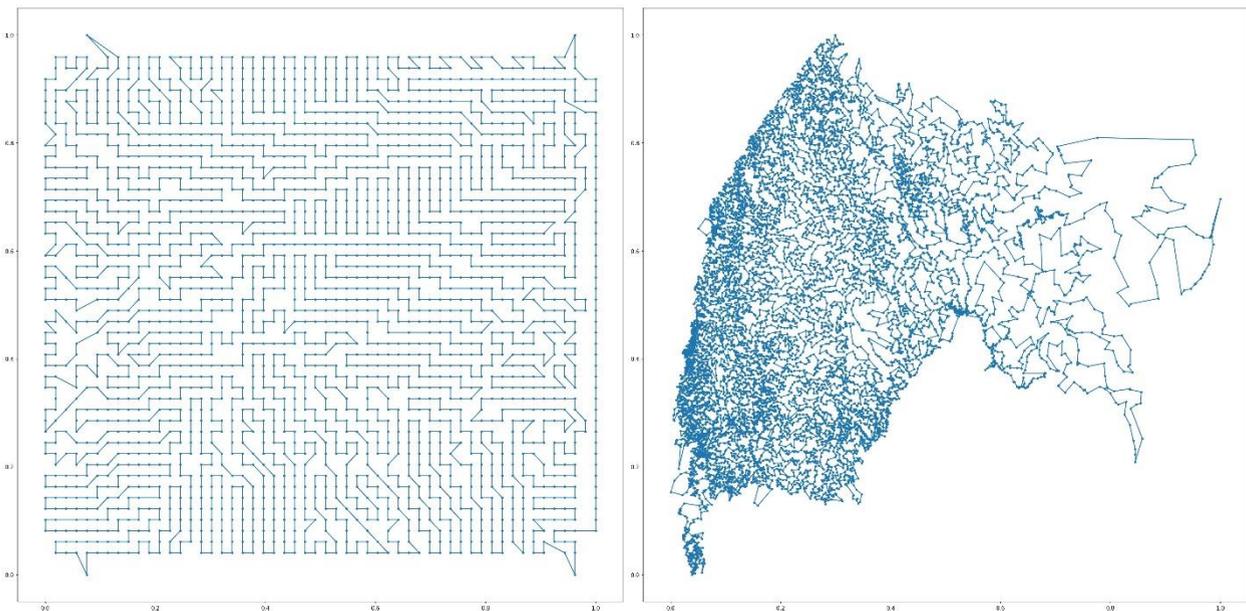

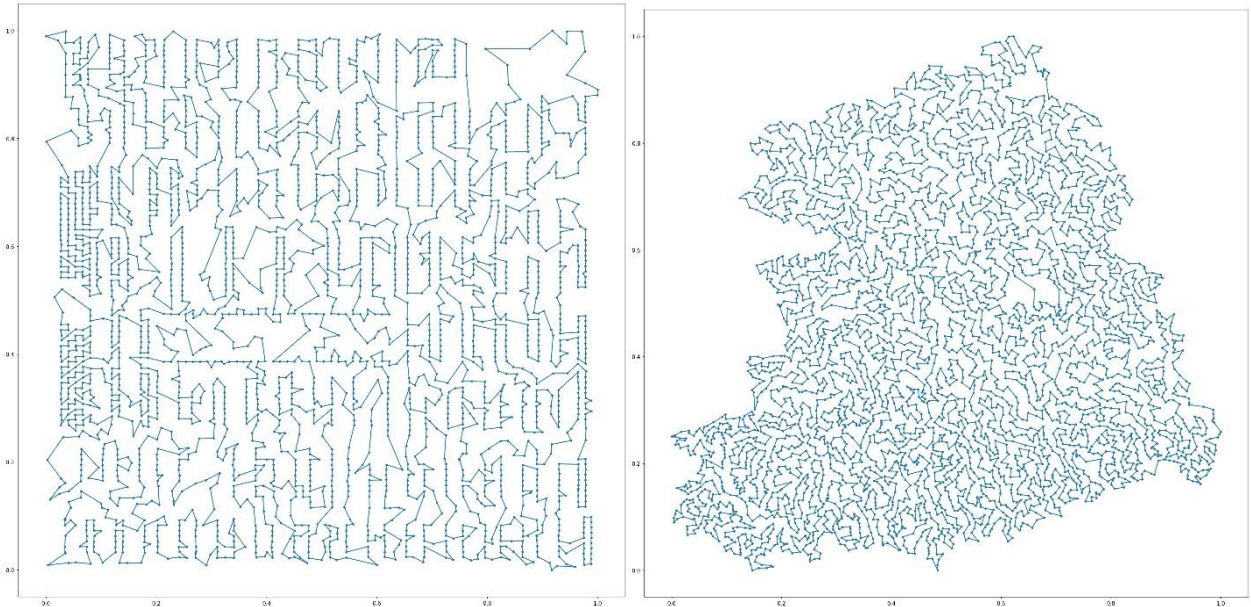

**Figure 8 Sample tours for benchmark instances**

## VI. CONCLUSION AND FUTURE WORK

In this work, a hybrid pointer network (HPN) is proposed for tackling both small-scale and large-scale problems. We demonstrate that the hybrid concept with a graph-based method is successful in improving the model performance for both scales. We used REINFORCE with Rollout Baseline to train our model. Our results show that our model outperforms the traditional graph pointer network with a significant margin, resulting in improved model generalization. In the future, we will attempt to find a robust architecture to improve the quality and the time of solutions for large-scale problems, resulting in better model generalization. We also want to tackle combinatorial problems with constraints, which will be an important direction for future study.